\documentclass[11pt]{article}

\usepackage{acl}

\usepackage{times}
\usepackage{latexsym}

\usepackage[T1]{fontenc}

\usepackage[utf8]{inputenc}

\usepackage{microtype}

\usepackage{inconsolata}

\usepackage{graphicx}

\usepackage{booktabs}
\usepackage{multirow}
\usepackage{amsmath}
\usepackage{tcolorbox}
\usepackage{listings}
\usepackage{blindtext}
\usepackage{cleveref}

\title{Comprehensive Comparison of RAG Methods Across \\Multi-Domain Conversational QA}

\author{
 \textbf{Klejda Alushi},
 \textbf{Jan Strich},
 \textbf{Chris Biemann,
 \textbf{Martin Semmann}
 }\\
 Hub of Computing and Data Science (HCDS) \\
 University of Hamburg, Germany \\
 \\
 \small{
   \textbf{Correspondence: \texttt{\{first\_name\}.\{last\_name\}@uni-hamburg.de}}
 }
}

\begin{document}
\maketitle

\begin{abstract}
Conversational question answering increasingly relies on retrieval-augmented generation (RAG) to ground large language models (LLMs) in external knowledge. Yet, most existing studies evaluate RAG methods in isolation and primarily focus on single-turn settings. This paper addresses the lack of a systematic comparison of RAG methods for multi-turn conversational QA, where dialogue history, coreference, and shifting user intent substantially complicate retrieval. We present a comprehensive empirical study of vanilla and advanced RAG methods across eight diverse conversational QA datasets spanning multiple domains. Using a unified experimental setup, we evaluate retrieval quality and answer generation using generator and retrieval metrics, and analyze how performance evolves across conversation turns. Our results show that robust yet straightforward methods, such as reranking, hybrid BM25, and HyDE, consistently outperform vanilla RAG.
In contrast, several advanced techniques fail to yield gains and can even degrade performance below the No-RAG baseline. We further demonstrate that dataset characteristics and dialogue length strongly influence retrieval effectiveness, explaining why no single RAG strategy dominates across settings. 
Overall, our findings indicate that effective conversational RAG depends less on method complexity than on alignment between the retrieval strategy and the dataset structure.
We publish the code used.\footnote{\href{https://github.com/Klejda-A/exp-rag.git}{GitHub Repository}}

\end{abstract}

\section{Introduction} \label{ch:1-introduction}
Conversational search, the task of satisfying information needs through multi-turn dialogue, has gained significant traction due to recent advances in LLMs~\citep{mo_survey_2025}.
The field is shifting from traditional keyword-based queries to conversational search, characterized by multi-turn natural-language interactions that capture complex and evolving information needs~\citep{mo_survey_2025, prayitno_conversational_2025}. 
To support these interactions, RAG has emerged as the de facto standard~\citep{lewis_retrievalaugmented_2020, huang_survey_2024, nikishina_creating_2025}. By retrieving external evidence from vector databases, RAG mitigates hallucinations and ensures responses are factually grounded and up-to-date~\citep{shuster_retrieval_2021, sahoo_comprehensive_2024}.

\begin{figure}[!t]
\begin{center}
\includegraphics[width=1\columnwidth]{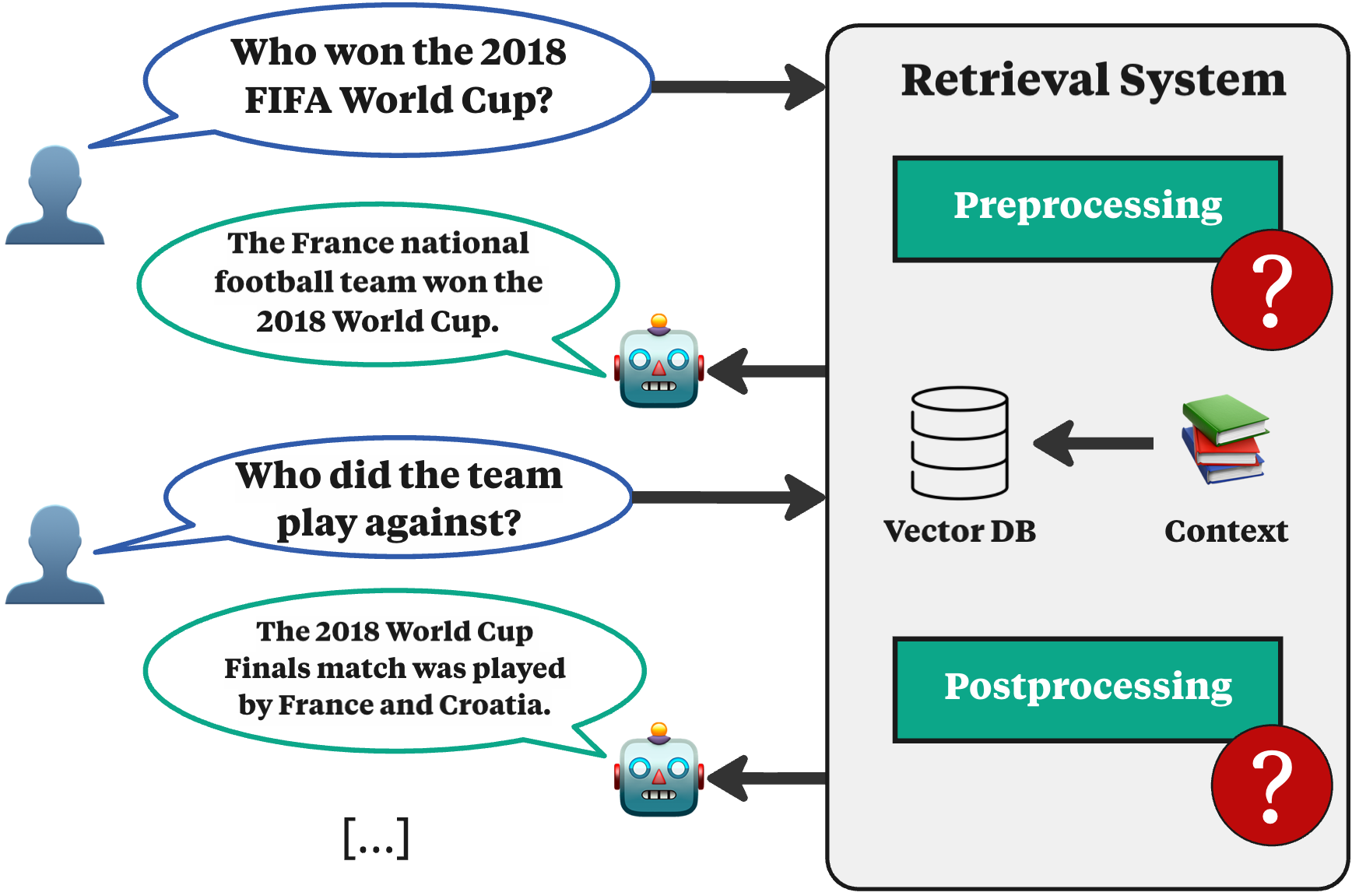}
\caption{Conversational Search Problem. One sample from the INSCIT dataset~\cite{wu_inscit_2023}. }
\label{fig:overview}
\end{center}
\end{figure}

While RAG is well established for single-turn Question Answering (QA), effectively integrating external knowledge into multi-turn conversations introduces significant complexity. In this setting, the system must maintain context across the dialogue history, resolving coreference and handling implicit queries when information is omitted (e.g., ellipsis). Consequently, retrieval effectiveness can vary widely depending on the system's ability to track dialogue history, resolve ambiguity, and adapt to shifting user intent across turns~\citep{saharoy_evidence_2025, chang_mainrag_2025, zhang_dhrag_2025}.
Although the literature reports a rapid evolution of advanced RAG architectures~\citep{gao_retrievalaugmented_2023, huang_survey_2024} and optimization strategies~\citep{gao_complement_2021, gao_precise_2023}, these methods are typically evaluated in isolation~\cite{yu_evaluation_2025}.

Currently, the field lacks a comprehensive overview of RAG strategies for conversational settings. Existing studies often utilize only vanilla RAG~\cite{liu_chatqa_2024, xu_chatqa_2025}, lacking SOTA retrieval metrics, limiting reproducibility, and practical insights for RAG in production systems. 
Furthermore, the interplay between retrieval performance and the depth of the conversation, specifically how performance degrades or changes as the dialogue progresses, remains underexplored.

To address this gap, we present an empirical analysis of RAG methods for conversational QA.
Our contributions are as follows: 
\begin{itemize}
    \item We provide a unified comparison of vanilla RAG and \textbf{six advanced RAG methods} under a reliable evaluation on \textbf{eight conversational QA datasets.}
    \item We further analyze the influence of the \textbf{position of the conversational turn} on retrieval performance. 
\end{itemize}

\section{Related Work}

\paragraph{LLMs and Conversational QA.}
Adapting LLMs for Conversational QA is typically achieved via fine-tuning a pre-trained model or by incorporating external context via RAG~\cite{dhabalia_comparative_2025}. Instruction tuning, which aligns pre-trained LLMs with conversational instructions, has become a foundational approach~\cite{zhang_instruction_2025}. These models have been successfully adapted for diverse domains, ranging from general knowledge~\cite{yang_hotpotqa_2018, joshi_triviaqa_2017} to specialized fields such as medicine~\cite{li_chatdoctor_2023, prayitno_conversational_2025} and law~\cite{wu_study_2025}. Fine-tuning strategies often involve training on human-rewritten queries~\cite{mo_convgqr_2023} or converting multi-turn interactions into single-turn problems~\cite{ye_enhancing_2023}.

\paragraph{RAG and Conversational QA.}
Conversational QA benefits significantly from RAG~\cite{lewis_retrievalaugmented_2020}, which integrates additional context by retrieving semantically similar documents from a vector database. 
Recent work has enhanced this process by incorporating meta-information~\cite{saharoy_evidence_2025} and query rewriting to facilitate accurate generation~\cite{mo_convgqr_2023}. Further advancements include self-check mechanisms, in which the model assesses the correctness of its own answers~\cite{ye_boosting_2024}, and learning policies that determine when and what to retrieve~\cite{roy_learning_2024}.
Beyond these conversation-specific adaptations, the broader RAG landscape offers numerous state-of-the-art methodologies~\cite{gao_retrievalaugmented_2023, huang_survey_2024} that hold potential for this domain. However, existing research often focuses on the end-to-end performance of one method, rather than comparing promising methods from the literature~\cite{gao_complement_2021, tito_document_2021, gao_precise_2023}. Consequently, this work presents a comprehensive comparison of these advanced retrieval strategies within conversational QA.

\paragraph{Conversational QA Datasets.} \label{sec:conversational_qa}
Conversational QA has evolved from single-turn tasks (e.g., SQuAD~\cite{rajpurkar_squad_2016}) to complex multi-turn settings. While datasets such as HotPotQA~\cite{yang_hotpotqa_2018} and 2WikiMultiHopQA~\cite{ho_constructing_2020} emphasize multi-hop reasoning, contemporary benchmarks have shifted their focus toward 
retrieval and conversational fidelity. 
PopQA~\cite{mallen_when_2023} addresses long-tail knowledge retrieval, while ChatQA~\cite{liu_chatqa_2024} and ChatQA-2~\cite{xu_chatqa_2025} establish a new standard for conversational QA by evaluating models on their ability to reason over retrieved evidence within fluid, multi-turn dialogues. Building on this foundation, we leverage the ChatQA~\cite{liu_chatqa_2024} dataset to systematically analyze how distinct RAG strategies perform across different knowledge domains.

\section{RAG Methods} \label{sec:rag}

\paragraph{Without RAG.}
We establish two reference points: \textit{No RAG}, which sends queries directly to the LLM using only the conversation history, and \textit{Oracle Context}, which provides ground-truth contexts.
\textit{No RAG} measures the LLM's internal capabilities through pretraining and sets dataset-specific baselines, whereas \textit{Oracle Context }simulates perfect retrieval to define a ceiling for the generator. 

\paragraph{Basic RAG Methods.}
This category comprises methods that retrieve documents using standard embedding-based retrieval techniques. The \emph{Base RAG} approach follows the original RAG framework~\cite{lewis_retrievalaugmented_2020}, in which only the input query is embedded to retrieve the top-$k$ documents, which are then passed unchanged to the generator. As a purely lexical baseline, \emph{Standard BM25}~\cite{robertson_probabilistic_2009} ranks documents based on term-frequency and inverse document-frequency statistics, relying on keyword overlap between the query and documents. \emph{Hybrid BM25}~\cite{gao_complement_2021} combines sparse BM25 retrieval with dense vector retrieval, leveraging the complementary strengths of lexical and semantic matching to improve recall and relevance. Finally, the \emph{Reranker} method~\cite{glass-etal-2022-re2g} applies a cross-encoder after initial retrieval to reorder documents according to their significance in a shared embedding space.

\paragraph{Advanced RAG Methods.}
This category encompasses methods that enhance retrieval through either \emph{preprocessing} or \emph{postprocessing} strategies. Preprocessing methods modify the input query to improve retrieval quality. The \emph{HyDE} method~\cite{gao_precise_2023} generates hypothetical answers for each query and uses them as refined queries to retrieve more relevant documents. In contrast, \emph{Query Rewriting}~\cite{ye_enhancing_2023} reformulates the original query to better align with the target document distribution. Postprocessing methods operate on retrieved contexts to improve their usefulness for generation. \emph{Summarization} reduces contextual noise by condensing each retrieved document using an LLM, focusing on salient information. \emph{SumContext} applies a similar summarization step while retaining the original full documents for generation, aiming to reduce distractions while preserving content fidelity. Finally, the \emph{HyDE Reranker} performs post-retrieval reranking by leveraging hypothetical answer generation to reorder initially retrieved documents based on semantic alignment.


\section{Datasets}\label{sec:datasets}

For our evaluation, we use ChatRAG-Bench~\cite{liu_chatqa_2024}, a benchmark comprising 10 conversational QA datasets covering diverse topics and formats. We select eight of these subsets for our experiments, as detailed in Table~\ref{table:datasets}. We exclude HybridDial~\cite{nakamura_hybridialogue_2022} due to the lack of annotated ground-truth contexts, which renders accurate retriever evaluation infeasible. Additionally, we omit ConvFinQA~\cite{chen_convfinqa_2022} because the answers primarily consist of numerical results derived from simple arithmetic operations, making the F1 score an unreliable metric for performance. The remaining eight subsets contain question-context-answer triples, enabling the independent evaluation of both retriever and generator components. Data preprocessing involved normalizing datasets from Hugging Face to a standardized format: we serialized multi-turn dialogues into a linear text format and removed formatting artifacts from context passages.
We describe each dataset in detail below.



\begin{table*}[!ht]
\centering
\resizebox{\textwidth}{!}{
\begin{tabular}{l l r r| r r r r}
\toprule
\multirow{2}{*}{\textbf{Subset}}  
& \multirow{2}{*}{\textbf{Source}}  
& \multirow{2}{*}{\textbf{\# Contexts}} 
& \multirow{2}{*}{\textbf{\# QA Pairs}} 
& \multirow{2}{*}{\textbf{Ctx/Q Ratio}} 
& \multicolumn{3}{c}{\textbf{Avg. Tokens}} \\
\cmidrule(lr){6-8}
& & & & & Question & Answer & Context \\
\midrule
QuAC         & Wikipedia      & 26,315   & 7,350  & 3.58 & 8.81  & 19.91 & 511.42 \\
SQA          & Wikipedia      & 185      & 3,010  & 0.06 & 10.69 & 37.26 & 453.83 \\
QReCC        & Wikipedia      & 19,275   & 2,790  & 6.91 & 8.12  & 28.16 & 505.52 \\
TopiOCQA     & Wikipedia      & 169,231  & 2,510  & 67.42 & 9.12  & 17.30 & 97.12 \\
\hline
Doc2Dial     & Social Welfare & 1,238    & 3,940  & 0.31 & 12.52 & 22.77 & 350.38 \\
DoQA         & StackExchange  & 395      & 1,790  & 0.22 & 13.16 & 19.00 & 145.50 \\
\hline
CoQA         & Mixed          & 499      & 7,980  & 0.06 & 7.69  & 4.43  & 329.38 \\
INSCIT       & Mixed          & 29,497   & 502    & 58.76 & 12.23 & 45.32 & 101.17 \\
\midrule
\textbf{Total | W. Avg} & Multiple & \textbf{246,635} & \textbf{29,872} & 8.25 & 9.47 & 18.82 & 175.81 \\
\bottomrule
\end{tabular}
}
\caption{Summary of the QA datasets, including source, number of contexts and QA pairs, context-to-question ratio, and average token lengths for questions, answers, and contexts. Weighted averages are computed across all subsets, using the number of QA pairs and contexts. Avg. token count based on Llama 3.3 tokenizer.}
\label{table:datasets}
\end{table*}


\paragraph{Sequential Question Answering (SQA)}
SQA \cite{iyyer_searchbased_2017} is a conversational dataset that focuses on a QA dialogue regarding semi-structured, Wikipedia tables. The aim was to decompose long, complex questions into sequences of small, easy-to-answer sub-questions. WikiTableQuestions \cite{ho_constructing_2020} was used to source the original questions, filtering out those that required arithmetic or could not be answered directly from table cells.   

\paragraph{Question Answering in Context (QuAC)}
The QuAC \cite{choi_quac_2018} dataset focuses on QA-based conversations between a \textit{teacher} and a \textit{student} regarding a Wikipedia article about an entity. The student knows only the article title, whereas the \textit{teacher} has full access; instead of answering the question in free text, they can only reply with a text excerpt. The interaction continues until one of three outcomes is reached: 12 questions have been answered, two questions remain unanswered, or either the student or the teacher decides to end the dialogue.

\paragraph{Conversations Question Answering (CoQA)}
\citet{reddy_coqa_2019} introduced the CoQA dataset to include diverse data sources, such as children's literature, school exams, and news, as well as casual-sounding speech, in which questions often refer to the dialogue history and answers are direct and without explanation. Each question may have multiple correct answers to account for grammatical or formatting differences. The evaluation is performed between the generated answer and each reference answer, and only the reference with the highest F1 score is selected.

\paragraph{Domain-specific Question Answering (DoQA)}
DoQA \cite{campos_doqa_2020} is a dataset built upon a continuous dialogue between a \textit{user} and a \textit{domain expert} relating to a specific topic, in this case cooking, travel, or movie forums on Stack Exchange. DoQA aimed for a more natural conversation, based more on follow-up questions than clunky factoids, and, similar to CoQA, there are four correct answers for each question. 

\paragraph{Doc2Dial}
Doc2Dial \cite{feng_doc2dial_2020}, a dataset consisting of dialogues between a \textit{user} and an \textit{agent} on topics related to social welfare in the United States as found on \textit{ssa.gov} and \textit{va.gov}. To showcase both dialogue- and document-based contexts, the conversations were sorted into three different categories: \textit{D1}, containing multiple questions relating to the given context, \textit{D2}, in which the conversation revolves around one central inquiry with clarifying questions carried out by the agent, and \textit{D3}, with questions that are irrelevant to the context. 

\paragraph{Question Rewriting in Conversational Context (QReCC)}
The QReCC \cite{ye_enhancing_2023} dataset incorporates questions from pre-existing datasets, including QuAC, and includes information sourced from the Common Crawl and web searches. Question rewriting was also used to "fix" any inquiries that had references to the conversation history, thereby preserving the natural-sounding sentence structure while removing ambiguity. These methods involve \textit{replacing} pronouns with their explicit referent, \textit{inserting} the referenced entity into the query itself, and \textit{removing} any unnecessary words.

\paragraph{Topic switching in Open-domain Conversational Question Answering (TopiOCQA)}
TopiOCQA \cite{adlakha_topiocqa_2022a} focuses on topic switching during a free-form conversation between a \textit{questioner} and an \textit{answerer}. The \textit{answerer} is granted full access to links within the relevant Wikipedia article. In contrast, the \textit{questioner} may only view the metadata and adjust their inquiries accordingly. The questions were divided into general open-ended questions, inquiries about specific entities, and requests for further details. The answers were unrestricted and free-form, facilitating topic switches every 3-4 questions and enabling handling of changing contexts and long-term reasoning.

\paragraph{Information-Seeking Conversations with Mixed-Initiative Interactions (INSCIT)}
\citet{wu_inscit_2023} proposed INSCIT, an information-seeking dataset that would use a variety of interactions between a \textit{user} and an \textit{agent} to challenge hard-to-answer questions regarding Wikipedia passages. 
It aimed to provide answer structures, categorizing them into: \textit{Direct answers}, with the \textit{agent} providing what they believe is the correct answer, \textit{Relevant answers}, which inform the user of relevant information regarding the query,  \textit{Clarifications}, which prompt the user for further information, and \textit{No information}, if not relevant answer is found.

\begin{figure*}[tb]
  \centering
  \includegraphics[width=\linewidth]{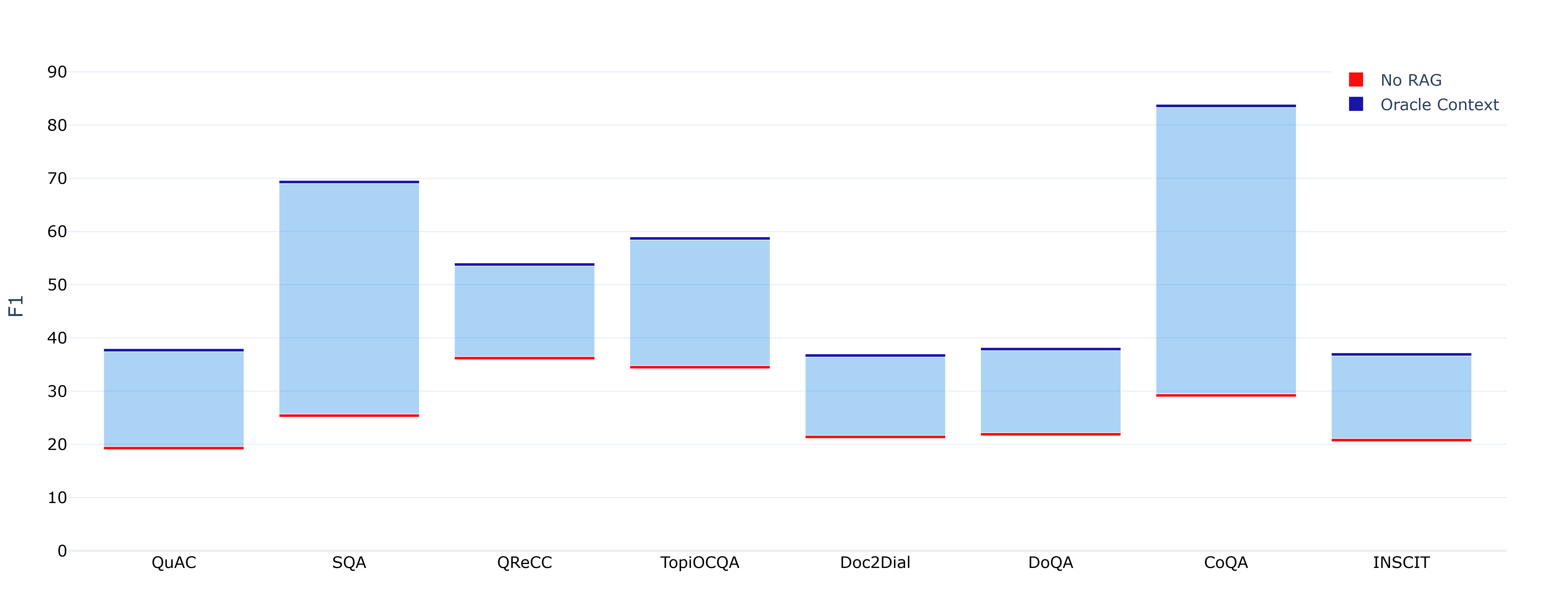}
  \caption[Min-Max F1 Range]{Theoretical minimum and maximum ranges of F1 that can be achieved with the LLM. Minimum is achieved by \textit{No RAG} method, which retrieves no contexts, whereas the maximum is achieved by the \textit{Oracle Context} method, which directly uses the gold label context.}\label{fig:f1_min_max_range}
\end{figure*}

\subsection{Dataset Statistics} \label{sec:statistics}

In total, our analysis encompasses over \textbf{29,000} QA pairs and more than \textbf{245,000 } distinct contexts across multiple domains. As shown in Table~\ref{table:datasets}, the datasets exhibit significant structural variation. The ratio of context tokens to query tokens varies considerably across the different subsets, whereas the average question length remains relatively balanced (7--12 tokens). In contrast, answer and context lengths exhibit substantial variance: average answer lengths range from 4 tokens (CoQA) to 45 tokens (INSCIT), whereas average context lengths range from approximately 100 to 500 tokens. 
Table \ref{table:datasets} presents the answer lengths, which vary considerably, ranging from the more concise answers in the CoQA dataset to the lengthier explanations in INSCIT, which were already accounted for in the system prompts.
While longer contexts risk containing more distracting content and reducing generator performance, shorter contexts can be disadvantageous if they do not provide sufficient context. 
A larger number of total contexts, as in the QuAC, INSCIT, and TopiOCQA datasets, could complicate or delay context retrieval. 

\subsection{Dataset Analysis} \label{sec:preliminary analysis}
To assess whether the datasets are suitable for analysis and whether they theoretically benefit from RAG, we conducted a pre-study in which we queried each dataset with the model using no information and with all information about the query.
We evaluate two control settings to disentangle pre-trained knowledge from the effect of context: \textit{No RAG}, which queries the LLM without external context, and \textit{Oracle Context}, which provides only the ground-truth context and serves as an upper bound, as shown in Figure~\ref{fig:f1_min_max_range}. 
We define the \textit{Oracle Context} setup as the upper bound of the model, given the generator's ability to answer the question using the golden context. 
In contrast, the lower bound is the LLM’s performance without any context, independent of the pre-trained model's knowledge.

For most datasets, the ceiling F1 remains below 40\%, with overall values around 15–20\%, indicating room for improvements through retriever methods alongside additional performance bottlenecks. CoQA and SQA exhibit wider F1 ranges, facilitating more precise comparisons between retrieval methods.
Furthermore, except for QReCC and TopiOCQA, the LLM displays comparable internal knowledge across datasets. Because LLMs are trained on public data, the \textit{No RAG} setting provides a valuable proxy for assessing overlap between the dataset and pre-training or fine-tuning data.

\begin{table*}[!th]
\centering
\resizebox{\textwidth}{!}{
\begin{tabular}{l l|cc|cc|cc|cc||cc|cc||cc|cc}
\toprule
\multirow{4}{*}{\textbf{RAG Method}} & \multirow{2}{*}{} 
  & \multicolumn{2}{c|}{\textbf{QuAC}}  
  & \multicolumn{2}{c|}{\textbf{SQA}}  
  & \multicolumn{2}{c|}{\textbf{QReCC}}  
  & \multicolumn{2}{c|}{\textbf{TopiOCQA}}  
  & \multicolumn{2}{c|}{\textbf{Doc2Dial}}  
  & \multicolumn{2}{c|}{\textbf{DoQA}}  
  & \multicolumn{2}{c|}{\textbf{CoQA}}  
  & \multicolumn{2}{c}{\textbf{INSCIT}} \\
  \cmidrule(lr){3-18} 
& & \multicolumn{8}{c|}{Wikipedia} & \multicolumn{2}{c|}{Social Welfare} & \multicolumn{2}{c|}{StackExchange} & \multicolumn{4}{c|}{Mixed} \\
\cmidrule(lr){3-4} \cmidrule(lr){5-6} \cmidrule(lr){7-8} \cmidrule(lr){9-10} 
\cmidrule(lr){11-12} \cmidrule(lr){13-14} \cmidrule(lr){15-16} \cmidrule(lr){17-18} 
& & MRR & F1 & MRR & F1 & MRR & F1 & MRR & F1 & MRR & F1 & MRR & F1 & MRR & F1 & MRR & F1\\
\midrule
\textit{No RAG} & &  - & 19.0 & - & 25.1 & - & 35.9 & - & 34.2 & - & 21.1 & - & 21.6 & - & 28.9 & - & 20.5 \\
\textit{Oracle Context} & & 100 & 38.0 & 100 & 69.6 & 100 & 54.1 & 100 & 59.0 & 100 & 37.0 & 100 & 38.2 & 100 & 83.9 & 100 & 37.2 \\
\hline
\textit{Vanilla RAG} & & 35.3 & 25.4 & 66.1 & 43.8 & 36.6 & 36.5 & 8.7 & 33.9 & 49.0 & 26.3 & 92.0 & 36.2 & 72.5 & 58.3 & 8.0 & 19.2 \\
\textit{Hybrid BM25} & & \textbf{44.6} & 27.5 & 53.8 & 45.6 & 37.4 & 37.4 & 9.4 & 34.0 & 51.0 & 27.7 & 84.8 & \textbf{36.9} & 76.6 & 67.3 & 8.1 & 19.1 \\
\textit{Reranker} & & 41.6 & \textbf{29.2} & \textbf{71.0} & \textbf{51.3} & 35.3 & 36.0 & 9.8 & 34.2 & 54.0 & \textbf{28.7} & \textbf{93.1} & 36.7 & 78.9 & \textbf{74.7} & 9.5 & 19.1 \\
\hline
\textit{Query Rewriting} & & 35.3 & 20.8 & 66.1 & 38.1 & 36.4 & 32.5 & 8.7 & 26.9 & 49.0 & 21.6 & 92.0 & 27.6 & 72.4 & 47.5 & 8.0 & 16.7 \\
\textit{HyDE} & & 42.0 & 26.2 & 65.3 & 38.0 & \textbf{49.0} & \textbf{42.2} & \textbf{25.1} & \textbf{43.5} & \textbf{57.5} & \textbf{28.7} & 90.9 & 35.4 & \textbf{86.6} & 71.3 & \textbf{25.2} & \textbf{25.9} \\
\textit{HyDE + Reranker} & & 30.5 & 25.4 & 61.6 & 38.4 & 37.5 & 37.0 & 14.6 & 37.6 & 48.0 & 26.4 & 85.2 & 34.7 & 58.2 & 53.0 & 13.9 & 22.0 \\
\textit{Summarization} & & 38.6 & 24.3 & 57.8 & 34.9 & 38.3 & 39.2 & 6.1 & 27.9 & 44.3 & 25.6 & 84.5 & 29.4 & 67.6 & 48.6 & 7.4 & 18.4 \\
\textit{SumContext} & & 37.7 & 26.3 & 57.7 & 35.1 & 40.0 & 37.1 & 6.7 & 27.5 & 44.9 & 26.7 & 85.3 & 35.4 & 68.0 & 62.7 & 7.4 & 18.5 \\
\bottomrule
\end{tabular}
}
\caption{Overall performance (MRR@5 and F1) of RAG methods on all eight conversational QA datasets. MRR@5 is used for retrieval performance and F1 for the generator. \textbf{Bold} values indicate the maximum for each column.}
\label{tab:rag_f1_mrr_final}
\end{table*}

\section{Evaluation and Results} \label{sec:data}

This section outlines the experiments we conducted to investigate the effect of RAG methods on multi-domain conversational QA. Firstly, we describe the experimental setups, prompt design, and evaluation metrics used in the assessment in ~\Cref{subsec:setup,subsec:promptdesign,subsec:metrics}. This is followed by discussing the results for the retriever and generator, and followed by the analysis of the relation of retriever and generator performance in~\Cref{subsec:retriever_analysis,subsec:generator_analysis,subsec:interaction}. We further analyze the effect of the conversation turn and discuss the results in~\Cref{sec:ablation_studies,ch:6-discussion}.

\subsection{Experimental Setup}\label{subsec:setup}
We evaluated all advanced RAG methods across all eight datasets using the EncouRAGe~\cite{strich_encourage_2025} library with the Llama 3 8B Instruct model \cite{grattafiori_llama_2024}, selected for its strong language understanding and manageable computational requirements. Additionally, the results of Gemma 3 27b~\cite{kamath_gemma_2025} were added in the camera-ready version in Appendix~\ref{apx:gemma3} and align with all results.
Key parameters were set to ensure reproducibility and efficiency: temperature = 0, maximum output length = 1000 tokens, and context length = 40,000 tokens. 
We conducted multiple runs for each method but observed only negligible differences across runs, so we reported results for only one run per method and dataset. 
Inference was performed on an NVIDIA RTX A6000 GPU with 48 GB of memory. EncouRAGe~\cite{strich_encourage_2025} facilitated dataset and RAG method management, integrating \texttt{vLLM} \cite{kwon_efficient_2023} for efficient batched inference and MLflow for experiment tracking. External contexts were stored in the Chroma vector database \cite{chromateam_chroma_2025} using Sentence Transformers \texttt{all-MiniLM-L6-v2}~\cite{reimers_making_2020} embeddings and Cosine Similarity for semantic relevance, ensuring efficient retrieval and evaluation across all methods and datasets.

\subsection{Prompt Design} \label{subsec:promptdesign}
Our prompt strategy relies on a consistent zero-shot template, following ChatQA~\cite{liu_chatqa_2024}. The general system prompt instructs the model to prioritize retrieved context and dialogue history, strictly avoiding reliance on internal knowledge to reduce hallucinations. Variations in the prompt were limited to formatting constraints (e.g., extraction vs. generation) to match specific dataset targets; the full set of dataset-specific prompt templates is provided in Appendix~\ref{apx:expanded_prompts}.

\subsection{Evaluation Metrics}\label{subsec:metrics}
For generators, we considered using F1, with F1 adopted as the primary metric to align with prior ChatRAG-Bench studies \cite{liu_chatqa_2024}. We selected the F1 Score from the SQuAD paper~\cite{rajpurkar_squad_2016} to balance token-level precision and recall, and, for datasets with multiple valid answers, we used the maximum score across references. Retriever performance was assessed using Recall@$k$, indicating whether the ground-truth context appears in the top-$k$ retrievals, and Mean Reciprocal Rank (MRR)~\cite{kantor_trec5_2000} for \textit{k} = 5, which emphasizes correct contexts ranked higher. These metrics together capture both the accuracy and ranking quality of retrieval, facilitating fair comparison across RAG methods.

\begin{figure*}[tb]
  \centering
  \includegraphics[width=\linewidth]{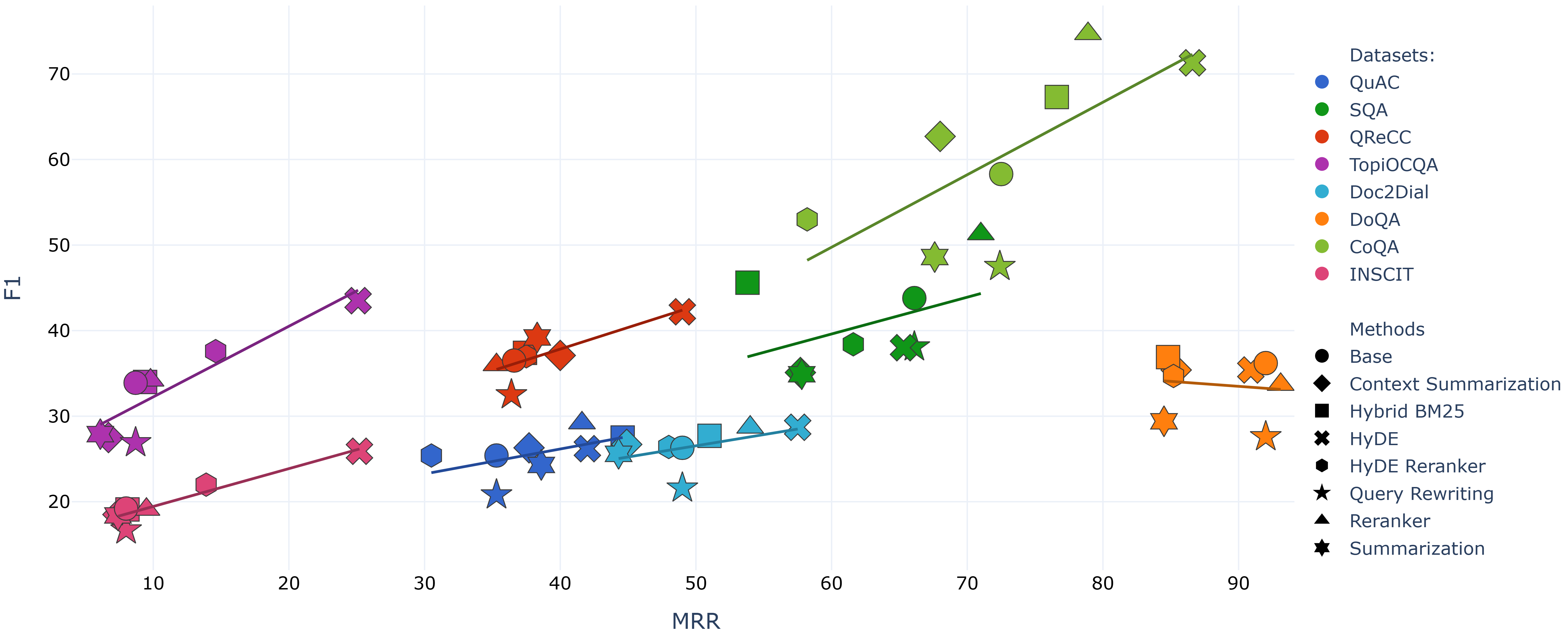}
  \caption[Distribution of F1 and MRR by method and dataset]{Relationship between retriever (MRR@\textit{5}) and generator (F1) performance for each dataset and method.}  \label{fig:scatter_f1_mrr_method_dataset}
\end{figure*}

\subsection{Retrieval Results}\label{subsec:retriever_analysis}
Table~\ref{tab:rag_f1_mrr_final} presents the results of the generator and retriever for each of the RAG methods. We also report Recall@\textit{1} and Recall@\textit{5} for each approach in Appendix~\ref{apx:retrieval}.
In terms of MRR@\textit{5} performance, for all datasets except for SQA~\cite{iyyer_searchbased_2017} and DoQA~\cite{campos_doqa_2020}, the combination of sparse and dense encoders in \textit{Hybrid BM25} leads to better results than \textit{Vanilla RAG}. This effect is also visible for the \textit{Reranker}.

Among the advanced RAG methods, \textit{HyDE} ranks clearly ahead, achieving the best performance on five of eight datasets. 
It is important to highlight that for INSCIT, \textit{HyDE} triples the performance of \textit{Vanilla RAG}.
The two summarization methods yielded poor retrievals, suggesting that summarization may remove crucial contextual information.
Dataset-wise, the MRR@\textit{5} values were highest in the DoQA and CoQA datasets and were significantly lower in INSCIT and TopiOCQA, which are analyzed further in Section \ref{sec:ablation_studies}.

\subsection{Generator Results}\label{subsec:generator_analysis}
Table~\ref{tab:rag_f1_mrr_final} shows that \textit{Hybrid BM25} consistently achieves slightly higher F1 scores than \textit{Vanilla RAG} across all datasets. Consistent with the retriever results, \textit{HyDE} emerges as the strongest approach, attaining the highest performance on four of the eight datasets and highlighting its effectiveness for conversational QA. This advantage is particularly evident on TopiOCQA, where \textit{HyDE} improves the F1 score by 9.6\% over \textit{Vanilla RAG}. In contrast, the performance of \textit{Query Rewriting} is highly dataset-dependent and falls substantially below \textit{No RAG} on the INSCIT, QReCC, and TopiOCQA datasets, where MRR scores are also below average. This may point to differences in conversational structure, such as larger topic shifts or longer dependency chains, that render these datasets less effective for this method. Finally, for summarization methods, incorporating the original conversational context yields only marginal improvements except for DoQA and CoQA.

\subsection{Relationship of Retriever and Generator}\label{subsec:interaction}


Figure~\ref{fig:scatter_f1_mrr_method_dataset} illustrates the relationship between retrieval and generation, with a fitted linear regression line summarizing the overall performance trend. Overall, F1 and MRR are positively correlated, indicating that stronger retrieval generally leads to higher answer quality, particularly on INSCIT, TopiOCQA, CoQA, and SQA, where \textit{HyDE} and \textit{HyDE Ranker} follow this trend. For CoQA and SQA, \textit{Reranker} appears to perform best, yielding the highest overall results on these datasets.

In contrast, QuAC and Doc2Dial show only marginal differences across methods, and the correlation is weaker. This disconnect is most pronounced for DoQA, where MRR exceeds 90\% but F1 remains below 40\%, highlighting that strong retrieval does not necessarily translate into strong generation. 

Furthermore, Spearman's  $\rho$ in Table~\ref{table:spearman} illustrates that most datasets have a relatively high correlation between the F1 and MRR values. The only exceptions being SQA and DoQA, which show weak or even negative correlations, suggesting that the two metrics capture different aspects of method performance.

These differences can be attributed to dataset specific problems, particularly relating to variations in answer formats. Answer conciseness is associated with overall high F1 scores, particularly when comparing CoQA's short responses against QReCC and INSCIT's longer, more in-depth answers, which are more challenging to match. 

\begin{table}[!htbp]
\centering
\begin{tabular}{l | l}
\toprule
\textbf{Subset} & \textbf{$\rho$} \\
\midrule
QuAC         & 0.620 \\
SQA          & 0.383 \\
QReCC        & 0.857 \\
TopiOCQA     & 0.874 \\
Doc2Dial     & 0.687 \\
DoQA         & -0.157 \\
CoQA         & 0.762 \\
INSCIT       & 0.788 \\
\bottomrule
\end{tabular}
\caption{Illustration of the Spearman's rank correlation coefficient ($\rho$), calculated for the F1 and MRR values for each dataset.}
\label{table:spearman}
\end{table}

The answer content can also heavily influence the performance, such as the social welfare answers of Doc2Dial, which rely on specific case details (user eligibility, personal details) and predetermined scripts, with many answers taking the form of follow-up questions requesting further information. Similarly, the forum-based DoQA dataset contains many informal responses in which respondents draw on personal knowledge rather than relying solely on the provided context. Additionally, the sensitive nature of certain DoQA queries leads to the generator refusing to respond, on the grounds that it cannot provide legal advice or discuss topics relating to violence or weaponry.

\begin{figure*}[tb]
  \centering
  \includegraphics[width=\linewidth]{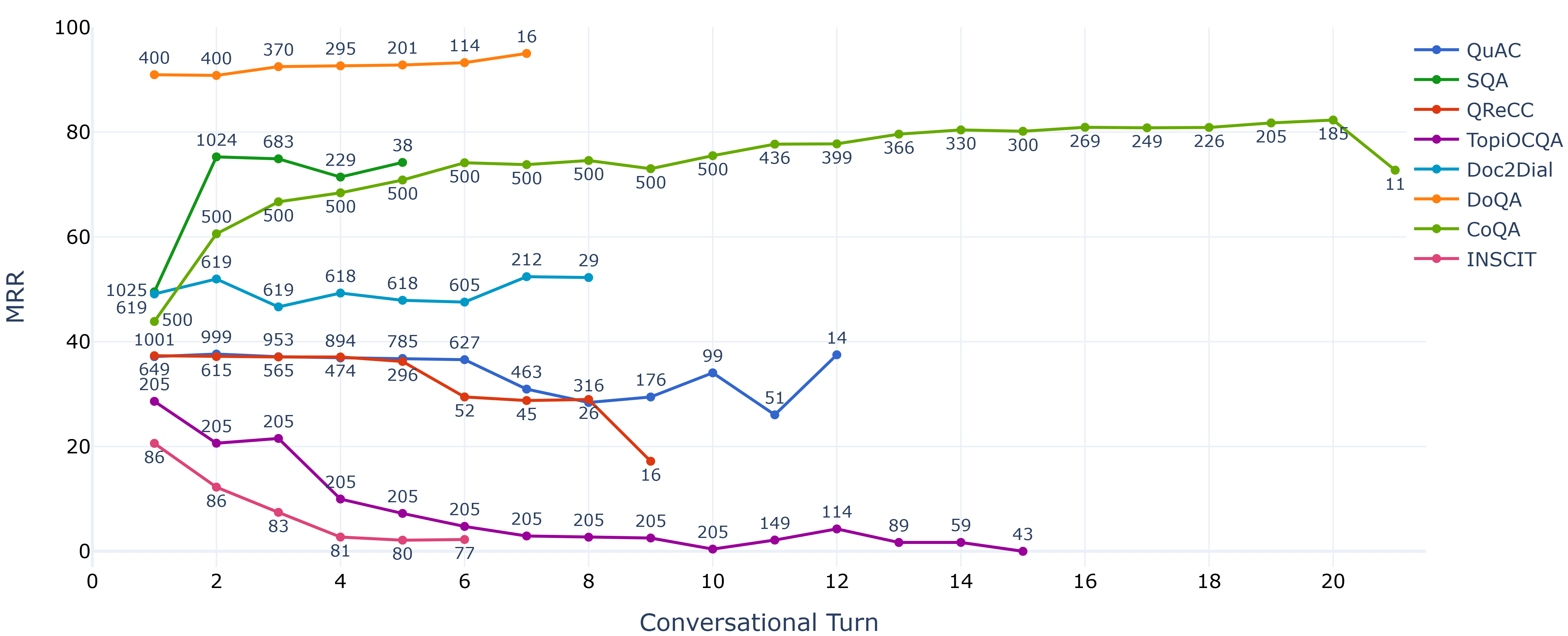}
  \caption[MRR development over conversation turns]{MRR performance across conversational turn for each dataset using \textit{Vanilla RAG}. The annotations indicate the number of samples per turn for each dataset.}\label{fig:conv_mrr_lim}
\end{figure*}

\subsection{Ablation Study} \label{sec:ablation_studies}
To further understand the effect of RAG on conversational QA, we examine retrieval performance across conversational turns in Figure~\ref{fig:conv_mrr_lim}. We compute MRR and F1 Score by turn position to identify trends in retrieval performance and present the F1 results in Appendix~\ref{apx:conv_f1}.
We find mixed results for all eight datasets. It is expected that INSCIT and TopiOCQA exhibit low performance that steadily decreases with the number of turns, primarily due to the high Ctx/Q ratio, as shown in Table~\ref{table:datasets}. This generally leads to low retrieval performance, and both datasets are designed to support topic and interaction entity switching.

In contrast, CoQA and SQA benefit from more context and improve performance with each turn, indicating that, when the context is consistent, more information leads to better retrieval performance. We found in addition that for QReCC, QuAC, DoQA, and Doc2Dial, all datasets seem to show no difference regarding the number of conversational turns til turn 5. There, we find that QReCC and QuAC show performance decreases that warrant further investigation in future work.

Overall, the results suggested that when the context is consistent, retrieving it from the entire conversation is beneficial; however, when questions are topic-switching or when other entities are interacting, this approach can decrease retrieval performance. We therefore recommend that calculating the similarity between queries and conversational histories can benefit retrieval performance.

\subsection{Discussion}\label{ch:6-discussion}%
This section examines whether conclusive findings were obtained and, if so, how they could inform future studies in Conversational QA or RAG.  We showed in our preliminary analysis that adding ground-truth context boosted performance by 15–50\%, providing a ceiling for RAG methods and a baseline for weaker ones. About half of the RAG methods performed on par with \textit{No RAG}, showing that inefficient pipelines either retrieve irrelevant contexts or fail to rank the ground-truth context highly enough.


As observed in the experiments, the results varied considerably across datasets and RAG methods, making it difficult to draw a unified conclusion. For F1, the top three methods were \textit{Reranker}~\cite{glass-etal-2022-re2g}, \textit{Hybrid BM25}~\cite{gao_complement_2021}, and \textit{HyDE}~\cite{gao_precise_2023}, respectively, indicating that \textit{Vanilla RAG}~\cite{lewis_retrievalaugmented_2020} was clearly outperformed across all datasets. Therefore, in one respect, the two advanced methods are recommended as superior alternatives in terms of performance. 
    
In contrast, it is essential to examine how the advanced RAG methods affect computational complexity and runtime overhead relative to \textit{Vanilla RAG.} To calculate relevance scores, \textit{Reranker} retrieves a larger number of candidate contexts, which are then passed through a cross-encoder. On the other hand, \textit{Hybrid BM25} adds a sparse retrieval function to the existing dense retrieval. Overall, although both methods are computationally simple, the experiment runs indicate that the double-retrieval process and score calculation of \textit{Hybrid BM25} result in slightly longer runtimes. 
While \textit{Vanilla RAG} does not achieve the highest overall performance, its high F1 scores and straightforward implementation make it a worthwhile baseline. Since the advanced methods do not deviate substantially from the baseline computationally, they also provide a valuable reference for estimating the potential performance of other advanced RAG methods. 

When examining the impact of dataset characteristics, the preliminary analysis revealed a substantial difference in the model's internal knowledge of specific dataset topics or formats. This is evident in the F1 range difference between Doc2Dial~\cite{feng_doc2dial_2020}, centered on specialized, open-ended questions in comparison to the factoid-style questions of CoQA~\cite{reddy_coqa_2019}. The total number of contexts significantly affected the retriever's performance by reducing the likelihood of finding the correct context.

\section{Conclusion} \label{sec:conclusion}

In this paper, we presented a systematic empirical study of vanilla and advanced RAG methods across eight multi-turn conversational QA datasets. Our evaluation, which accounted for both retriever effectiveness and generator quality, revealed that robust yet straightforward techniques, such as \textit{Reranker} and \textit{Hybrid BM25},  consistently outperform \textit{Vanilla RAG} across all evaluated domains. Among the advanced techniques studied, the HyDE method proved the most effective for enhancing retrieval performance, whereas the \textit{Reranker} approach was most successful at improving final answer quality. For future research, the results highlight that performance improvements can be achieved without inflating the computational complexity, emphasizing the need to prioritize retrieval strategies over resource-intensive scaling. 

Furthermore, our analysis of conversational turns demonstrated that the impact of dialogue depth varies substantially across datasets, reflecting their distinct structures. While some settings benefit from the accumulation of consistent dialogue history, others suffer from performance degradation as the conversation progresses, particularly when handling topic switching or shifts in user intent. These results suggest that the effectiveness of conversational RAG is determined less by the inherent complexity of the retrieval method than by the strategic alignment between the retrieval strategy and the dataset's specific structural characteristics. We conclude that future advances in the field should prioritize this alignment to ensure that external knowledge is integrated accurately and efficiently into multi-turn dialogues.



\section*{Limitations}

\paragraph{Methodological and Dataset Heterogeneity.}
This study was hindered by the wide variety of RAG methods and datasets, which required extensive, dataset-specific preprocessing due to differing prompt formats, answer structures, and context representations. This heterogeneity increased experimental complexity and limited the depth of analysis across all method–dataset combinations. Future work could mitigate this issue by focusing on a smaller, more representative subset of methods and datasets, especially given the redundancy among many RAG enhancements.

\paragraph{Retrieval Challenges from Large and Fragmented Contexts.}
Another limitation arises from the large number of contexts per query, which makes retrieving the ground-truth passage difficult, particularly for TopiOCQA, QuAC, and INSCIT. This issue could be addressed by grouping contexts during pre-processing using metadata such as titles or by adopting iterative retrieval or generation strategies that increase the likelihood of retrieving relevant evidence and producing accurate answers.

\section*{Acknowledgement}
This work is supported by the Genial4KMU project, Universität Hamburg, funded by BMBF (grant no. 16IS24044B).

\bibliography{clean}

\clearpage
\appendix

\section{Expanded Dataset Prompts}\label{apx:expanded_prompts}
The system prompts are designed to instruct the LLM on how best to answer the question and to emphasize the focus that should be placed on the previous conversation history and contexts provided, and if they do not provide the answer then the LLM should indicate as such, instead of relying on internal information. A segment of the prompt is used to guide the LLM on how the answer should be formatted whether it be multiple sentences or short phrases.
\begin{figure}[!ht]
    \centering
    \small
\begin{tcolorbox}[
  colback=white,
        colframe=black,
        arc=0pt,
        boxrule=0.5pt,
        left=10pt,
        right=10pt,
        top=5pt,
        bottom=5pt,
        width=\linewidth,
        ]
\begin{verbatim}
CoQA: You are a helpful assistant who will try 
to answer the following question to the best of 
your abilities. Use only on the given context 
and conversation history and do not use any 
assumptions or external information. Make the 
answers as direct as possible without using any 
redundant information and without using full 
sentences. Indicate if you cannot find the 
answer based on the context.

Doc2Dial: You are a helpful assistant. Answer 
the question strictly based on the given 
context. Do not use prior knowledge, make 
assumptions, or introduce any information not 
present in the context. If the answer is 
clearly stated, respond in a complete and 
concise sentence. If the context does not 
provide enough information, respond with a 
relevant follow-up question to clarify the 
user's intent.

DoQA: You are a helpful assistant trying to 
answer the questions to the best of your 
abilities. Use only the given context to answer 
the question. and do not use any assumptions or 
external information. Keep your answer 
relevant, direct and in one sentence. Do not 
explain the background, context or reasoning 
behind the answer. Do not refer to the context 
in your response. Indicate if you cannot find 
the answer based on the context.

INSCIT: You are a helpful assistant. Answer the 
question strictly based on the given context. 
Do not use prior knowledge, make assumptions, 
or include any information not present in the 
context. Do not refer to the context in your 
response. If the answer is not available, say 
so clearly. Respond in one full and 
complete sentence.

QReCC: You are a helpful assistant. Answer the 
question strictly based on the given context. 
Do not use prior knowledge, make assumptions, 
or introduce any information not present in 
the context. If the answer is not available, 
clearly state that. Respond in a single, clear, 
and complete sentence whenever possible.
\end{verbatim}
\end{tcolorbox}
\end{figure}

\begin{figure}[!ht]
    \centering
    \small
\begin{tcolorbox}[
  colback=white,
        colframe=black,
        arc=0pt,
        boxrule=0.5pt,
        left=10pt,
        right=10pt,
        top=5pt,
        bottom=5pt,
        width=\linewidth,
        ]
\begin{verbatim}
INSCIT: You are a helpful assistant. Answer the 
question strictly based on the given context. 
Do not use prior knowledge, make assumptions, 
or include any information not present in the 
context. Do not refer to the context in your 
response. If the answer is not available, say 
so clearly. Respond in one full and 
complete sentence.

QReCC: You are a helpful assistant. Answer the 
question strictly based on the given context. 
Do not use prior knowledge, make assumptions, 
or introduce any information not present in 
the context. If the answer is not available, 
clearly state that. Respond in a single, clear, 
and complete sentence whenever possible.

QuAC: You are a helpful assistant who will try 
to answer the following question to the best of 
your abilities. Use only the given context and 
conversation history and do not use any 
assumptions or external information. Keep your 
answer short, direct and in one sentence. Do 
not explain the background, context or 
reasoning behind the answer. Indicate if you 
cannot find the answer based on the context.

SQA: You are a helpful assistant. Use only the 
given table and conversation history to answer 
the question. Do not rely on outside knowledge 
or make assumptions. Return the exact answer 
from the table. Use brief phrases or values and 
no full sentences.

TopiOCQA: You are a helpful assistant who will 
try to answer the following question to the best 
of your abilities. Use only the given context 
and conversation history and do not use any 
assumptions or external information. Make the 
answers as direct as possible without using any 
redundant information and without using full 
sentences. Indicate if you cannot find the 
answer based on the context.
\end{verbatim}
\end{tcolorbox}
\captionsetup{font=small}
    \caption[System dataset prompts]{List of the system prompts used for each dataset.}
    \label{fig:generation-prompt-1}
\end{figure}

\onecolumn
\section{Performance of Gemma 3 27b}\label{apx:gemma3}

\begin{table*}[!th]
\centering
\resizebox{\textwidth}{!}{
\begin{tabular}{l l|cc|cc|cc|cc||cc|cc||cc|cc}
\toprule
\multirow{4}{*}{\textbf{RAG Method}} & \multirow{2}{*}{} 
  & \multicolumn{2}{c|}{\textbf{QuAC}}  
  & \multicolumn{2}{c|}{\textbf{SQA}}  
  & \multicolumn{2}{c|}{\textbf{QReCC}}  
  & \multicolumn{2}{c|}{\textbf{TopiOCQA}}  
  & \multicolumn{2}{c|}{\textbf{Doc2Dial}}  
  & \multicolumn{2}{c|}{\textbf{DoQA}}  
  & \multicolumn{2}{c|}{\textbf{CoQA}}  
  & \multicolumn{2}{c}{\textbf{INSCIT}} \\
\cmidrule(lr){3-18}
& & \multicolumn{8}{c|}{Wikipedia} 
  & \multicolumn{2}{c|}{Social Welfare} 
  & \multicolumn{2}{c|}{StackExchange} 
  & \multicolumn{4}{c|}{Mixed} \\
\cmidrule(lr){3-4} \cmidrule(lr){5-6} \cmidrule(lr){7-8} \cmidrule(lr){9-10}
\cmidrule(lr){11-12} \cmidrule(lr){13-14} \cmidrule(lr){15-16} \cmidrule(lr){17-18}
& & MRR & F1 & MRR & F1 & MRR & F1 & MRR & F1 & MRR & F1 & MRR & F1 & MRR & F1 & MRR & F1 \\
\midrule
\textit{No RAG} & & - & 21.8 & - & 30.1 & - & 31.5 & - & 25.2 & - & 21.1 & - & 28.5 & - & 17.2 & - & 13.8 \\
\textit{Oracle Context} & & 100 & 47.0 & 100 & 78.6 & 100 & 55.1 & 100 & 61.8 & 100 & 42.7 & 100.0 & 52.9 & 100 & 83.2 & 100 & 36.6 \\
\hline
\textit{Vanilla RAG} & & 35.3 & 30.4 & 66.1 & 51.0 & 36.6 & 31.5 & 8.7 & 25.2 & 49.0 & 28.6 & \textbf{92.0} & 50.8 & 72.5 & 57.3 & 8.0 & 14.7 \\
\textit{Hybrid BM25} & & \textbf{44.7} & 33.2 & 54.0 & 52.8 & 37.3 & 32.9 & 9.3 & 25.8 & 51.3 & 30.7 & 84.8 & 51.2 & 76.4 & 66.4 & 8.1 & 14.7 \\
\textit{Reranker} & & 43.7 & \textbf{36.3} & \textbf{72.6} & \textbf{58.1} & 34.9 & 29.9 & 9.7 & 26.7 & 55.4 & 31.3 & 93.6 & \textbf{51.0} & \textbf{81.3} & \textbf{75.5} & 9.6 & 15.3 \\
\hline
\textit{Query Rewriting} & & 35.4 & 24.8 & 66.1 & 46.7 & 36.5 & 33.4 & 8.7 & 19.3 & 49.0 & 25.5 & \textbf{92.0} & 37.8 & 72.6 & 51.6 & 8.0 & 14.2 \\
\textit{HyDE} & & 31.5 & 29.2 & 63.5 & 56.2 & \textbf{47.7} & \textbf{47.0} & \textbf{30.5} & \textbf{46.5} & \textbf{56.3} & \textbf{35.8} & 89.0 & 45.7 & 68.0 & 64.7 & \textbf{29.4} & \textbf{26.9} \\
\textit{HyDE + Reranker} & & 24.8 & 29.0 & 60.6 & 58.0 & 38.3 & 42.8 & 16.3 & 37.5 & 46.2 & 33.8 & 82.2 & 47.5 & 46.2 & 54.1 & 15.3 & 20.1 \\
\textit{Summarization} & & 37.8 & 24.8 & 63.5 & 55.7 & 39.4 & 34.6 & 7.5 & 29.1 & 34.6 & 23.6 & 85.3 & 33.2 & 70.9 & 38.9 & 6.7 & 16.4 \\
\textit{SumContext} & & 37.7 & 31.6 & 63.6 & 57.1 & 38.8 & 43.6 & 7.5 & 29.6 & 34.3 & 31.6 & 85.8 & 47.5 & 71.6 & 69.2 & 7.0 & 17.0 \\
\bottomrule
\end{tabular}
}
\caption{Overall performance (MRR@5 and F1) of RAG methods on all eight conversational QA datasets using Gemma 3 27b~\cite{kamath_gemma_2025}. MRR@5 is used for retrieval performance, and F1 for the generator. \textbf{Bold} values indicate the maximum for each column.}
\label{tab:gemma3_27b_rag}
\end{table*}

\section{Recall Retrieval Performance}\label{apx:retrieval}

\begin{table*}[!th]
\centering
\resizebox{\textwidth}{!}{
\begin{tabular}{l l|cc|cc|cc|cc||cc|cc||cc|cc}
\toprule
\multirow{4}{*}{\textbf{RAG Method}} & \multirow{2}{*}{} 
  & \multicolumn{2}{c|}{\textbf{QuAC}}  
  & \multicolumn{2}{c|}{\textbf{SQA}}  
  & \multicolumn{2}{c|}{\textbf{QReCC}}  
  & \multicolumn{2}{c|}{\textbf{TopiOCQA}}  
  & \multicolumn{2}{c|}{\textbf{Doc2Dial}}  
  & \multicolumn{2}{c|}{\textbf{DoQA}}  
  & \multicolumn{2}{c|}{\textbf{CoQA}}  
  & \multicolumn{2}{c}{\textbf{INSCIT}} \\
  \cmidrule(lr){3-18} 
& & \multicolumn{8}{c|}{Wikipedia} & \multicolumn{2}{c|}{Social Welfare} & \multicolumn{2}{c|}{StackExchange} & \multicolumn{4}{c|}{Mixed} \\
\cmidrule(lr){3-4} \cmidrule(lr){5-6} \cmidrule(lr){7-8} \cmidrule(lr){9-10} 
\cmidrule(lr){11-12} \cmidrule(lr){13-14} \cmidrule(lr){15-16} \cmidrule(lr){17-18} 
& & R@1 & R@5 & R@1 & R@5 & R@1 & R@5 & R@1 & R@5 & R@1 & R@5 & R@1 & R@5 & R@1 & R@5 & R@1 & R@5 \\
\midrule
\textit{Vanilla RAG} & & 24.9 & 52.6 & 58.2 & 79.2 & 25.0 & 56.0 & 5.4 & 14.2 & 36.0 & 70.1 & 88.3 & 97.2 & 66.2 & 81.7 & 3.8 & 15.7 \\
\textit{Hybrid BM25} & & 28.7 & \textbf{77.1} & 43.2 & 76.3 & 25.1 & 59.9 & 5.7 & 16.3 & 35.1 & \textbf{78.4} & 74.8 & \textbf{98.4} & 62.4 & \textbf{96.4} & 3.8 & 16.1 \\
\textit{Reranker} & & 28.4 & 63.4 & \textbf{61.7} & \textbf{86.0} & 24.6 & 53.2 & 6.6 & 15.4 & 40.4 & 75.1 & \textbf{90.1} & 97.1 & \textbf{72.3} & 88.9 & 6.0 & 15.5 \\
\hline
\textit{Query Rewriting} & & 24.9 & 52.6 & 58.2 & 79.2 & 25.0 & 56.1 & 5.4 & 14.2 & 36.0 & 70.1 & 88.3 & 97.2 & 66.3 & 81.7 & 3.8 & 15.7 \\
\textit{HyDE} & & \textbf{30.4} & 60.7 & 56.6 & 78.8 & \textbf{35.4} & \textbf{71.9} & \textbf{18.4} & \textbf{37.4} & \textbf{44.2} & \textbf{78.4} & 87.7 & 95.6 & 83.4 & 90.5 & \textbf{16.3} & \textbf{40.8} \\
\textit{HyDE + Reranker} & & 17.7 & 54.7 & 52.0 & 77.2 & 28.0 & 52.5 & 10.9 & 20.6 & 35.7 & 67.8 & 72.5 & 86.7 & 52.7 & 67.4 & 9.0 & 21.5 \\
\textit{Summarization} & & 27.7 & 56.6 & 49.2 & 72.3 & 26.8 & 58.9 & 4.0 & 10.2 & 31.4 & 65.1 & 80.6 & 92.1 & 58.7 & 79.9 & 4.0 & 13.3 \\
\textit{SumContext} & & 26.8 & 55.4 & 48.0 & 73.4 & 28.2 & 60.4 & 4.1 & 11.4 & 32.4 & 65.6 & 80.2 & 92.0 & 60.3 & 80.0 & 4.5 & 13.2 \\
\bottomrule
\end{tabular}
}
\caption{Overall performance (R@1 and R@5) of RAG methods on all eight conversational QA datasets. \textbf{Bold} values indicate the maximum for each column.}
\label{tab:rag_f1_final}
\end{table*}

\clearpage
\section{F1 Performance Across Conversational Turns}\label{apx:conv_f1}

\begin{figure*}[ht]
  \centering
  \includegraphics[width=1\linewidth]{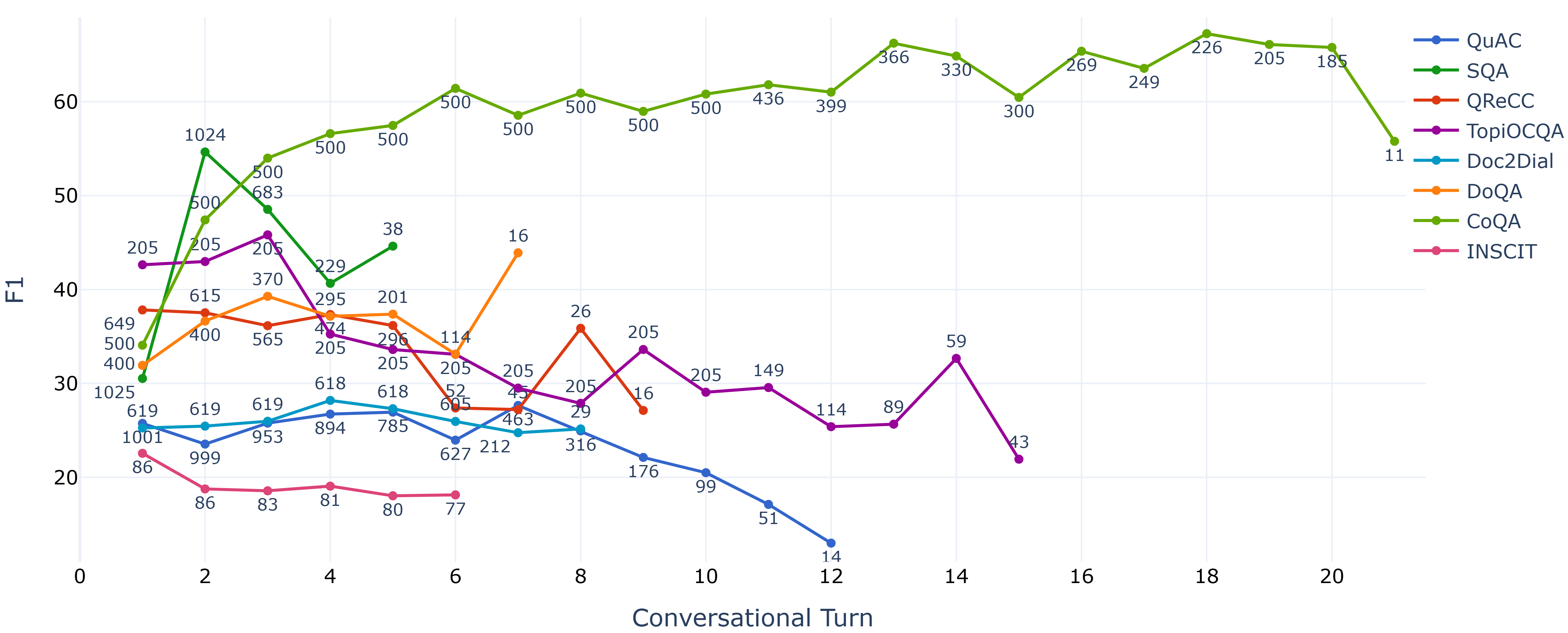}
  \caption[F1 development over conversation turns]{F1 performance across conversational turn for each dataset using \textit{Vanilla RAG}.}\label{fig:conv_f1_lim}
\end{figure*}

\end{document}